\documentclass{bmvc2k}

\usepackage{bm}
\usepackage{amsmath}
\usepackage{amssymb}
\usepackage{amsthm}
\usepackage{multirow}
\usepackage{array}
\usepackage{amsfonts}
\usepackage{physics}
\usepackage{graphbox}
\usepackage{booktabs}
\usepackage{subscript} 
\usepackage[ruled,vlined]{algorithm2e}
\usepackage{colortbl}
\definecolor{green1}{rgb}{0.4, 0.65, 0.3}
\definecolor{blue}{rgb}{0.21,0.49,0.74}
\usepackage{tabularx}
\newcolumntype{C}{>{\centering\arraybackslash}X}

\title{Making Rotation Averaging Fast and Robust with Anisotropic Coordinate Descent}
\addauthor{\quad\quad Yaroslava Lochman\vspace{-10pt}}{}{1}
\addauthor{\quad\quad Carl Olsson\vspace{-10pt}}{}{2}
\addauthor{\quad\quad Christopher Zach\vspace{-10pt}}{}{1}
\addinstitution{Chalmers University of Technology}
\addinstitution{Lund University}
\runninghead{Lochman et al.}{Fast and Robust Rotation Averaging  with ACD}

\def\Tr{\operatorname{tr}}

\newcolumntype{H}{>{\setbox0=\hbox\bgroup}c<{\egroup}@{}}
\def\m#1{\ensuremath{\mathtt{#1}}}

\def\R{\mathbb{R}}

\def\mI{{\m I}}

\def\mR{{\m R}}
\def\mH{{\m H}}
\def\mM{{\m M}}
\def\mN{{\m N}}

\def\tr{^\top}
\def\pinv{^\mathsf{\dagger}}

\DeclareMathOperator{\projectSOthree}{project_{SO(3)}}
\DeclareMathOperator{\SOthree}{SO(3)}

\def\allbutk{\tilde{k}}
\def\kth{\ensuremath{k^\text{th}}\xspace}

\newcommand{\teenytiny}{\fontsize{5}{5.6}\selectfont}
\newcommand{\ttiny}{\fontsize{4}{4.6}\selectfont}

\def\eg{\emph{e.g}\bmvaOneDot}

\def\etal{\emph{et al}\bmvaOneDot}
\def\ie{\emph{i.e}\bmvaOneDot}

\newcolumntype{V}{!{\vrule width 0pt}}

\SetKwInput{Require}{Require}

\begin{document}
\maketitle

\begin{abstract}
Anisotropic rotation averaging has recently been explored as a natural extension of respective isotropic methods. In the anisotropic formulation, uncertainties of the estimated relative rotations---obtained via standard two-view optimization---are propagated to the optimization of absolute rotations. The resulting semidefinite relaxations are able to recover global minima but scale poorly with the problem size. Local methods are fast and also admit robust estimation but are sensitive to initialization. They usually employ minimum spanning trees and therefore suffer from drift accumulation and can get trapped in poor local minima. In this paper, we attempt to bridge the gap between optimality, robustness and efficiency of anisotropic rotation averaging. We analyze a family of block coordinate descent methods initially proposed to optimize the standard chordal distances, and derive a much simpler formulation and an anisotropic extension obtaining a fast general solver. We integrate this solver into the extended anisotropic large-scale robust rotation averaging pipeline. The resulting algorithm achieves state-of-the-art performance on public structure-from-motion datasets. Project page: \href{https://ylochman.github.io/acd}{https://ylochman.github.io/acd}
\end{abstract}

\section{Introduction}
Rotation averaging is an important building block in global~\cite{moulon2012,moulon2017,pan2024} and hybrid~\cite{cui2017hsfm,chen2021hybrid} structure-from-motion (SfM) pipelines. The goal is to find absolute camera rotations that are globally aligned with the pre-estimated (via two-view optimization) noisy pairwise relative rotations.
Traditional methods for rotation averaging use isotropic error measures~\cite{dellaert2020,eriksson2021,parra2021,govindu2004,wilson2020,wilson2016}, where deviations between the observed and predicted relative rotations are penalized equally in all directions. A common approach is to minimize the so-called chordal distance~\cite{dellaert2020,parra2021,eriksson2021}, which is the least squares distance between the rotation matrices.
Since observed relative rotations in SfM are typically not direct measurements, but rather the result of an estimation from image data, covariance/precision matrices describing the certainty of the estimated rotation in all directions, can be extracted. 

It is well known~\cite{Zhang2023,olsson2025} that objective functions based on reprojection error in two-view-relative-pose problems can be insensitive to changes in certain directions, making the exact location of the best rotation hard to pinpoint. 
Since it is desirable to propagate such information to the later stages of the reconstruction pipeline, several recent works have looked into integrating uncertainties, through anisotropic objective functions, into rotation averaging. Zhang~\etal~\cite{Zhang2023} propose an approach to optimize the distances induced by two-view Hessians. Their results illustrate that rotation averaging can be made much more accurate than the previous isotropic versions, achieving camera poses that are closer to ground truth by taking uncertainties into account. 
The approach is based on local refinement of angle-axis representations of rotations.
To achieve a global method, Olsson~\etal~\cite{olsson2025} show how to integrate anisotropic cost with chordal distances into semidefinite programs (SDP)~\cite{vandenberghe1996semidefinite}. Solving such problems is equivalent to finding the maximum likelihood estimation (MLE) under an anisotropic Langevin distribution, where the anisotropic terms approximate uncertainties in the input rotations obtained as solutions of two-view SfM problems. The resulting SDP turns out to be tight significantly less often than the corresponding isotropic variant.
To address that, a stronger convex relaxation is proposed in~\cite{olsson2025} that is generally capable of finding the global optimum and leads to more accurate solutions than the isotropic counterpart. On the downside, the SDP formulation has a large number of additional constraints, and therefore the employed general purpose solvers suffer from poor scalability, restricting the applicability of the method to small-scale reconstruction problems.

\begin{figure}[t]
\centering
\caption{Comparison of the solutions to the isotropic versus anisotropic rotation averaging (RA) using the synthetic loop scene (left). Comparison between the state-of-the-art methods and the proposed \textbf{ACD} algorithm on the ``boulders'' dataset from ETH3D~\cite{schops2017multi} (right).}
\includegraphics[height=0.28\textwidth]{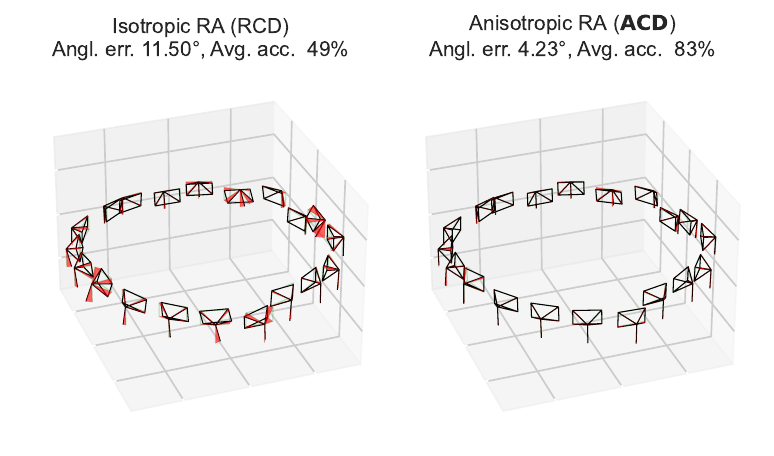}
\includegraphics[height=0.28\textwidth]{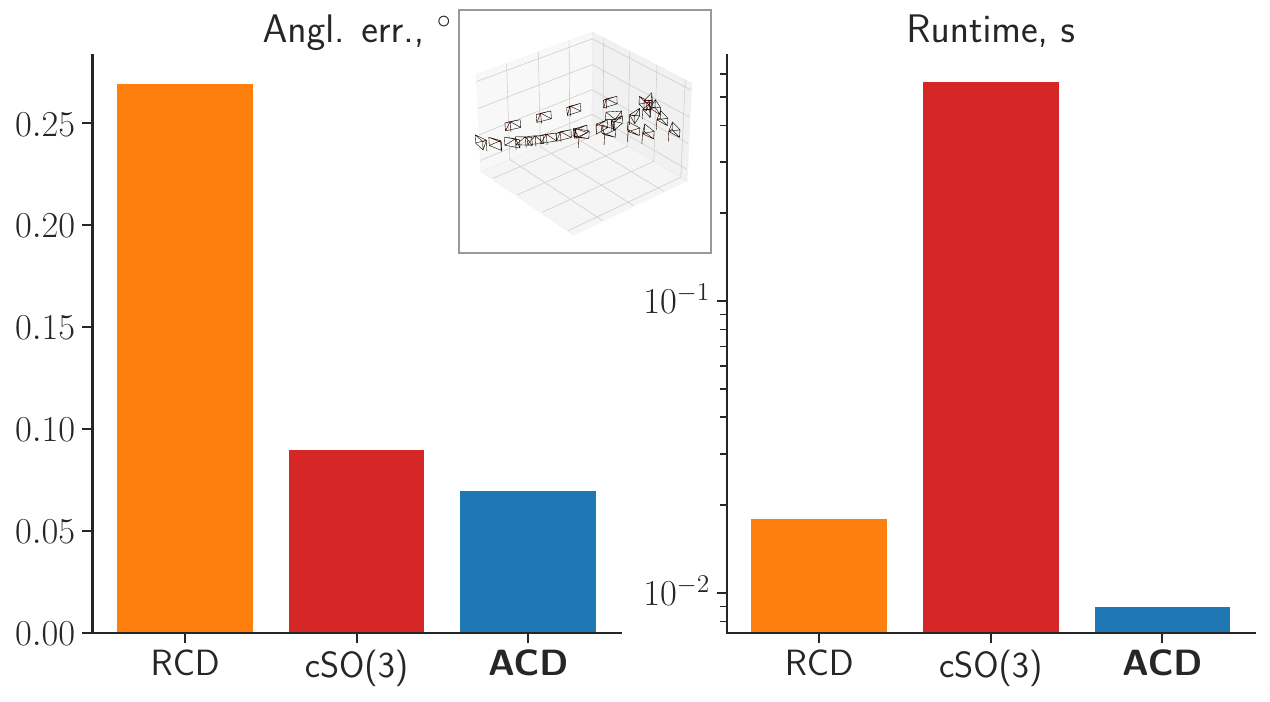}\\
\label{fig:teaser}
\end{figure}

The design of efficient algorithms therefore has been a key challenge for making large scale certifiable anisotropic rotation averaging feasible. In this paper, we try to address this problem. We start by looking at the block coordinate descent algorithms \cite{eriksson2018,parra2021} used for optimizing chordal distance objective in isotropic rotation averaging, and
extend this framework to incorporate uncertainties. In summary, our contributions are:
(1) we propose a simple and efficient algorithm for anisotropic rotation averaging called anisotropic coordinate descent (ACD);
(2) we analyze ACD and show empirically that it has a high success rate from different initializations;
(3) we propose an efficient pipeline for accurate, fast and robust anisotropic rotation averaging that achieves  state-of-the-art performance on the tested structure-from-motion datasets.
In Figure~\ref{fig:teaser}, we illustrate the advantages of ACD in accuracy and/or runtime in comparison with the recent methods of Parra~\etal~\cite{parra2021} (RCD) and Olsson~\etal~\cite{olsson2025} (cSO(3)).

\section{Related Work}
A common formulation of rotation averaging starts with minimizing the sum of chordal distances, $\sum_{i,j} \|\tilde \mR_{ij} - \mR_j \mR_i\tr\|^2_F$, subject to $\m R_i \in \SOthree$~\cite{arie2012global,fredriksson2012,hartley2013,eriksson2018,dellaert2020,eriksson2021,moreira2021rotation}. Under the isotropic Langevin noise~\cite{chiuso2008} assumption, this optimization problem corresponds to MLE. Martinec~\etal~\cite{martinec2007robust} relaxes the rotation constraints and proposes a sparse homogeneous linear system to obtain an approximate solution. 
Convex relaxations using Lagrangian duality are able to retain orthogonality constraints and have been shown to admit certifiably correct solutions \cite{arie2012global,fredriksson2012,eriksson2018,eriksson2021}.
While the resulting SDP programs can be solved with off-the-shelf solvers, much more efficient dedicated algorithms have also been developed for isotropic rotation averaging. 
Moreira~\etal~\cite{moreira2021rotation} designed a primal-dual sequence algorithm that is very efficient. Another highly efficient method was proposed by Rosen~\etal~\cite{rosen2019se} for pose graph optimization and further developed by Dellaert~\etal~\cite{dellaert2020}. It uses a rank-restricted semidefinite relaxation equivalent to the unconstrained Riemannian optimization on a product of Stiefel manifolds. The employed solution method is known as the Riemannian staircase algorithm~\cite{boumal2015riemannian}. Spectral relaxations lack optimality certificates but generally have comparable accuracy to SDP formulations for isotropic problems \cite{singer2011,arie2012global,doherty2022performance}.

To address outliers among the input data, robustified chordal distances have been employed \cite{wang2013exact,carlone2018convex}. Wang~\etal~\cite{wang2013exact} proposed a semidefinite relaxation for the least unsquared deviations (LUD) that also guarantees exact and stable recovery under a specific noise model. The resulting relaxation is solved using an alternating direction method~\cite{wang2013exact}. A similar approach is used in \cite{carlone2018convex} to derive convex relaxations and a posteriori guarantees for robust SE(2) synchronization with the LUD, $l_1$ and Huber loss.
Lie-algebraic rotation averaging~\cite{govindu2004} locally minimizes $\sum_{i,j}\|\text{log}(\tilde \mR_{ij} \mR_i \mR_j\tr)\|^2_F$. Chatterjee~\etal~\cite{chatterjee2013efficient,chatterjee2017} use the first-order distance approximation to obtain an iterative refinement scheme. To make the approach robust, they propose an iterative $l_1$ minimization followed by iteratively reweighted least squares.

Local optimizers rely on accurate initialization to ensure convergence to a good optimum. Exact inference along the edges of a minimum spanning tree (MST) of the camera graph is a common initialization strategy~\cite{chatterjee2013efficient,chatterjee2017,parra2021,Zhang2023}. Since MSTs lack redundancy, measurement noise can not be suppressed resulting in accumulated drift that may lead to failure in large problems~\cite{cornelis2004drift}.
Several works propose different heuristics to robustify the initial estimation of absolute rotations. 
Govindu~\etal~\cite{govindu2006robustness} samples multiple MSTs from the view-graph using RANSAC~\cite{fischler1981random}, while
Zach~\etal~\cite{zach2010disambiguating} proposes a Bayesian inference approach to identify likely perceptual aliases
from sampled cycles of the view graph.
Lee~\etal~\cite{lee2022hara} constructs a hierarchical spanning tree based on the consistency and reliability of the triplet constraints.

\section{Anisotropic Rotation Averaging}
\paragraph{Preliminaries} All mono-spaced matrices, \eg $\mR$, are $3\times3$ matrices. All bold matrices, \eg $\mathbf R$, are formed by stacking the respective mono-spaced matrices either vertically ($\mathbf R$, $\mathbf S$) or both vertically and horizontally ($\mathbf N$, $\tilde{\mathbf R}$, $\mathbf X$, $\mathbf B$). Let $\cal G = (\cal V, \cal E)$ be the view graph with $n$ vertices in $\cal V$ representing absolute rotations $\{\mR_1,...,\mR_n\}$, and edges $\cal E$ representing estimated two-view relations, \ie $(i,j) \in \cal E$ if an estimate of the relative rotation $\tilde \mR_{ij}$ exists. 

We remark that a common scenario is that $\tilde \mR_{ij}$ is obtained from an optimization of a two-view SfM problem. Assuming that $\tilde \mR_{ij}$ has been optimized to a minimum, the local behavior of the objective function is described by a Hessian, which can also be seen as the inverse of the covariance matrix. Hence such estimates usually come with uncertainty estimates, in this case describing the change in reprojection error of the two-view objective when deviating from the estimate $\tilde{\m R}_{ij}$. 

We follow the formulation from \cite{olsson2025} and seek to optimize anisotropic chordal distances
\begin{align}
    \min_{\mathbf R \in \SOthree^n} -\sum_{(i,j) \in \cal E} \left\langle \mM_{ij} \tilde \mR_{ij}, \mR_j \mR_i\tr \right\rangle,
    \label{eq:ARA_objective}
\end{align}
where $\mM_{ij}=\tfrac12 \Tr(\mH_{ij}) \mI - \mH_{ij}$, and $\mH_{ij}$ are Hessians of the optimized two-view objective functions. 
Further, we will equivalently write the objective in \eqref{eq:ARA_objective} as $-\left\langle\mathbf N, \mathbf R \mathbf R\tr\right\rangle$, where $\mathbf N$ is a symmetric block matrix containing the blocks $\mN_{ij} = \mM_{ij} \tilde \mR_{ij}$ for $(i,j) \in \cal E$\footnote{Note that $\mM_{ji} \tilde \mR_{ji} = (\mM_{ij} \tilde \mR_{ij})\tr$. For $i<j$,  $\mN_{ij} = (\mM_{ij} \tilde \mR_{ij})\tr$ and $\mN_{ji} = \mM_{ij} \tilde \mR_{ij}$.}, and zeros for $(i,j) \notin \cal E$. An isotropic version of $\mathbf N$ is $\tilde{\mathbf R}$ containing the blocks $\tilde \mR_{ij}$.

\subsection{Block coordinate descent methods} 
To derive an algorithm able to solve \eqref{eq:ARA_objective}, we take inspiration from the block coordinate descent methods~\cite{eriksson2018,parra2021} tackling the isotropic objective. The isotropic SDP formulation analyzed in~\cite{eriksson2021} is as follows,
\begin{align}
\min_{\mathbf X \succeq 0} -\left\langle\tilde{\mathbf R}, \mathbf X\right\rangle 
\quad\text{s.t. }
\m X_{ii} = \mI.
\end{align}
Under strong duality the optimal matrix $\mathbf X$ can be factorized into $\mathbf X = \mathbf R \mathbf R^T$ giving a certifiably optimal solution $\mathbf R$ to \eqref{eq:ARA_objective}.
A sub-problem is created by fixing all blocks except for the $\kth$ block-column/row of $\mathbf X$, and re-ordering the block-columns and respective block-rows in $\tilde{\mathbf R} $ and $\mathbf X$ such that the $\kth$ block-column/row becomes the first one
\begin{align}
\begin{split}
\min_{\mathbf S} & -\langle \mathbf W, \mathbf S\rangle
\quad
\text{s.t. }
\begin{pmatrix} \mI & \mathbf S\tr\\ \mathbf S & \mathbf B \end{pmatrix} \succeq 0,
\end{split}
\label{eq:BCD}
\end{align}
where $\mathbf W = \tilde{\mathbf R}_{\allbutk,k} \in \R^{3(n-1)\times3}$ is the \kth block-column of $\tilde{\mathbf R}$ excluding the \kth block-row, $\mathbf S \in \R^{3(n-1)\times3}$ is the block-column matrix of unknowns, $\mathbf B$ contains all fixed blocks. The problem in \eqref{eq:BCD} has a closed-form solution $\mathbf S^* = -\mathbf B \mathbf W ((\mathbf W\tr \mathbf B \mathbf W)^{1/2})\pinv$ which constitutes a single iteration of the block coordinate descent method in~\cite{eriksson2018}.

Instead of maintaining the $3n\times 3n$ matrix of unknowns $\mathbf X$, the rotation coordinate descent (RCD) algorithm~\cite{parra2021} only keeps the ``effective'' $3n\times 3$ matrix $\mathbf R$, motivated by the fact that the solution $\mathbf X$ admits factorization $\mathbf R \mathbf R\tr$. However, as a result, RCD does not increase the rank of the sought matrix $\mathbf X$ above $3$ at any iteration, and the method therefore reaches a local minimum over the set of rank-$3$ matrices.
Hence, the solution returned by RCD is not necessarily a global solution of the underlying semidefinite program.
Moreover, RCD is practically initialized with rotations, which, as we found, limits the solution of each iteration to ${\SOthree}^n$. To prove this, we need to show that for any $3\times3$ matrix $\m A$ the following holds
\begin{align}
    \m A\tr \left(\left(\m A \m A \tr\right)^{1/2}\right)^\dagger \in {\SOthree}.
\end{align}
This is easy to see if we let $\m A = \m U \m Q \m V \tr$ be the SVD of $\m A$. Then $\left(\m A \m A\tr\right)^{1/2} = \m U \m Q \m U \tr$ and $\m A\tr \left(\left(\m A \m A \tr\right)^{1/2}\right)^\dagger = \m V \m Q \m U \tr \m U \m Q^{-1} \m U \tr = \m V \m U \tr$.
Consequently, if $\mathbf R \in {\SOthree}^n$ then
\begin{align}
    \mathbf R \mathbf R\tr \mathbf W \left(\left(\mathbf W \tr \mathbf R \mathbf R\tr \mathbf W\right)^{1/2}\right)^\dagger \in {\SOthree}^n\label{eq:RCD_iteration}.
\end{align}
The expression in \eqref{eq:RCD_iteration} is the key operation of an RCD iteration (see Algorithm 2 in~\cite{parra2021}), and it restricts the search space to $\SOthree^n$. This, however, does not appear to be an obstacle in obtaining accurate reconstructions~\cite{parra2021} which we also confirm experimentally. We therefore make use of this finding as shown in the next paragraph.

\subsection{Anisotropic coordinate descent}
Now, we present how to leverage block coordinate descent in anisotropic rotation averaging. 
It was shown in \cite{olsson2025} that anisotropic costs require a stronger SDP of the form\vspace{-3pt}
\begin{equation}
    \min_{\mathbf X \succeq 0} -\left\langle \mathbf N, \mathbf X\right\rangle
    \quad\text{s.t. }
    \m X_{ii} = \mI
    \text{ and }
    \m X_{ij} \in 
\operatorname{convhull}(\SOthree).
\end{equation}
The additional constraint $\operatorname{convhull}(\SOthree)$ turns out to be significant. 
For both real and synthetic cases tested in \cite{olsson2025} 
optimal solutions could only be extracted when using the extra constraints. In contrast, the SDP turned out not to be tight when the convex hull constraints were omitted. 
Note that while $\operatorname{convhull}(\SOthree)$ is used in the SDP to achieve convexity, the optimal solution will still have $\m X_{ij} \in \SOthree$.
Based on these findings and the analysis in the previous paragraph, we propose to incorporate $\SOthree$ membership as a constraint which leads to a much simpler formulation.
We therefore consider the following sub-problem\vspace{-3pt}
\begin{align}
\begin{split}
\min_{\mathbf S \in \SOthree^{n-1}} & -\langle \mathbf N_{\allbutk,k}, \mathbf S\rangle
\quad
\text{s.t. }
\begin{pmatrix} \mI & \mathbf S\tr\\ \mathbf S & \mathbf B \end{pmatrix} \succeq 0,
\end{split}
\label{eq:ACD_problem}
\end{align}
where $\mathbf N_{\allbutk,k} \in \R^{3(n-1)\times3}$ is the \kth block-column of $\mathbf N$ excluding the \kth block-row. We note that, in addition to allowing anisotropic weights in $\mathbf N$, the problem in \eqref{eq:ACD_problem} is different from \eqref{eq:BCD} due to the added $\SOthree$ constraints. Using the generalized Schur complement (e.g.~\cite{burns1974generalized,zhang2006schur}), we rewrite the PSD constraint as $\mI - \mathbf S\tr \mathbf B\pinv  \mathbf S \succeq 0$ and $(\mI - \mathbf B \mathbf B\pinv)  \mathbf S = 0$.
Matrix $\mathbf B$ can be factorized as $\mathbf B = \mathbf R_{\allbutk} \mathbf R_{\allbutk}\tr$, where $\mathbf R_{\allbutk}$ is the  column-wise stacked rotations excluding the \kth one.
Similarly, $\mathbf S$ can be decomposed via $\mathbf S = \mathbf R_{\allbutk} \mR_k\tr$.
Using the properties of rotations, $\mathbf R_{\allbutk}\pinv = \mathbf R_{\allbutk}\tr$ and $\mathbf R_{\allbutk}\tr \mathbf R_{\allbutk} = \mI$, we rewrite the constraints as follows\vspace{-3pt}
\begin{align}
\begin{split}
& \mI - (\mathbf R_{\allbutk}\tr \mathbf S)\tr \mathbf R_{\allbutk}\tr  \mathbf S \succeq 0,
\qquad
(\mI - \mathbf R_{\allbutk} \mathbf R_{\allbutk}\tr) \mathbf S = 0.
\end{split}
\end{align}
Let $\m A = \mathbf R_{\allbutk}\tr  \mathbf S \in \R^{3\times3}$. From $(\mI - \mathbf R_{\allbutk} \mathbf R_{\allbutk}\tr) \mathbf S = 0$ we deduce $\mathbf S = \mathbf R_{\allbutk} \m A$. On the other hand, inserting $\mathbf S = \mathbf R_{\allbutk}~ \mR_k\tr$ yields $\m A = \mR_k\tr$, and $\m A$ is simply the sought-after (inverted) absolute rotation.
This implies that by reparametrizing \eqref{eq:ACD_problem} in terms of $\mR_k$ the PSD
constraint is automatically satisfied---hence redundant---, and we arrive at the following reformulation of~\eqref{eq:ACD_problem},
\begin{align}
\begin{split}
\min\nolimits_{\m \mR_k} & -\langle \mathbf N_k \tr \mathbf R, \m \mR_k\rangle
\quad
\text{s.t. }
\m \mR_k \in \SOthree,
\label{eq:ACD_reformulation}
\end{split}
\end{align}
where $\mathbf N_k \in \R^{3n\times3}$ is the \kth block-column of $\mathbf N$.\footnote{Note that now we use full block-columns in the coefficient matrix since $\m N_{k,k}$ is zeros.}
The solution to \eqref{eq:ACD_reformulation} is determined as
\begin{align}
    & \mR_k^* = \projectSOthree(\mathbf N_k \tr \mathbf R)
    \label{eq:ACD_solution_step}
\end{align}
where $\projectSOthree(\m M) = \m U \,\text{diag}(1,1,\text{det}(\m U \m V\tr)) \m V\tr$ with $[\m U, \sim, \m V] = \text{svd}(\m M)$.
Further, the solution of \eqref{eq:ACD_problem} is given by $\mathbf S^* = \mathbf R_{\allbutk}~ {\mR_k^*}\tr$.\footnote{We would like to point out that~\eqref{eq:ACD_solution_step} exactly coincides with minimizing~\eqref{eq:ARA_objective} w.r.t.\ $\m R_k$ for a fixed $k$.}

A simple block coordinate descent method for anisotropic rotation averaging \eqref{eq:ARA_objective} called anisotropic coordinate descent (ACD) is presented in Algorithm~\ref{alg:method}. Note that in isotropic setting ACD arrives at the same solution as RCD~\cite{parra2021} while performing much simpler computations. This is because at each iteration, RCD (1) does not leverage the properties of $\SOthree$ and instead performs additional matrix multiplications, and (2) rotates the world coordinate system so that $\mR_k$ is $\m I$, which is possible because of the gauge ambiguity, but is not necessary --- ACD directly modifies $\mR_k$ instead.

\begin{minipage}{0.92\textwidth}
\begin{algorithm}[H]
\small
\SetAlgoNoLine
\For{$t\in \{1, \dotsc\}$}{
\For{$k \in \text{shuffle}(\{1,\dots,n\})$}{
$\m R_i^{(t+1)} \gets \m R_i^{(t)} \quad (i\neq k);$
\qquad
$\m R_k^{(t+1)} \gets \projectSOthree(\mathbf N_k\tr {\mathbf R^{(t)}});$\\[1pt]
}}
\Return{$\mathbf R$}
\caption{ACD algorithm. Inputs: $\mathbf N$, $\mathbf R^{(0)}$.}
\label{alg:method}
\end{algorithm}
\end{minipage}\\[1em]
Due to the non-communicating blocks of constraints, the method converges to at least a stationary point~\cite{peng2023block}, often a good optimum\footnote{Similarly to RCD, it has a large convergence basin, but convergence to a global optimum is not guaranteed.} (see Section \ref{sec:ablation}). Finally, note that global SfM assumes an unordered image collection. 
Inspired by stochastic gradient descent, we shuffle the sequence of absolute rotations at each iteration $t$, and empirically verify the slightly improved performance.

\section{Experiments}
We evaluate the proposed and state-of-the-art rotation averaging algorithms on a number of synthetic and real datasets detailed below. We use the following metrics for evaluation: RMS angular error (Angl. err.), area under the cumulative rotation error curve at $n^\circ$ (AUC @ $n^\circ$), and average accuracy (AA) over the set of angular error thresholds $\{0.1^\circ,0.2^\circ\dots,19.9^\circ,20^\circ\}$. To remove gauge freedom, we post-multiply the estimated absolute rotations $\{R_i\}_i$ with $Q = \projectSOthree\left(\sum_i R_i\tr R^*_i\right)$, where $\{R^*_i\}_i$ are the ground-truth absolute rotations.

\subsection{Synthetic experiments}
\begin{figure}[h]
\centering
\caption{Synthetic results. Convergence of ACD from zero to a better solution as compared to RCD and chaining baseline on "loop" scenes (left). Error reduction of ACD versus RCD with the increasing perturbation scale in Hessian estimates on general scenes (right).}
\includegraphics[width=0.42\textwidth]{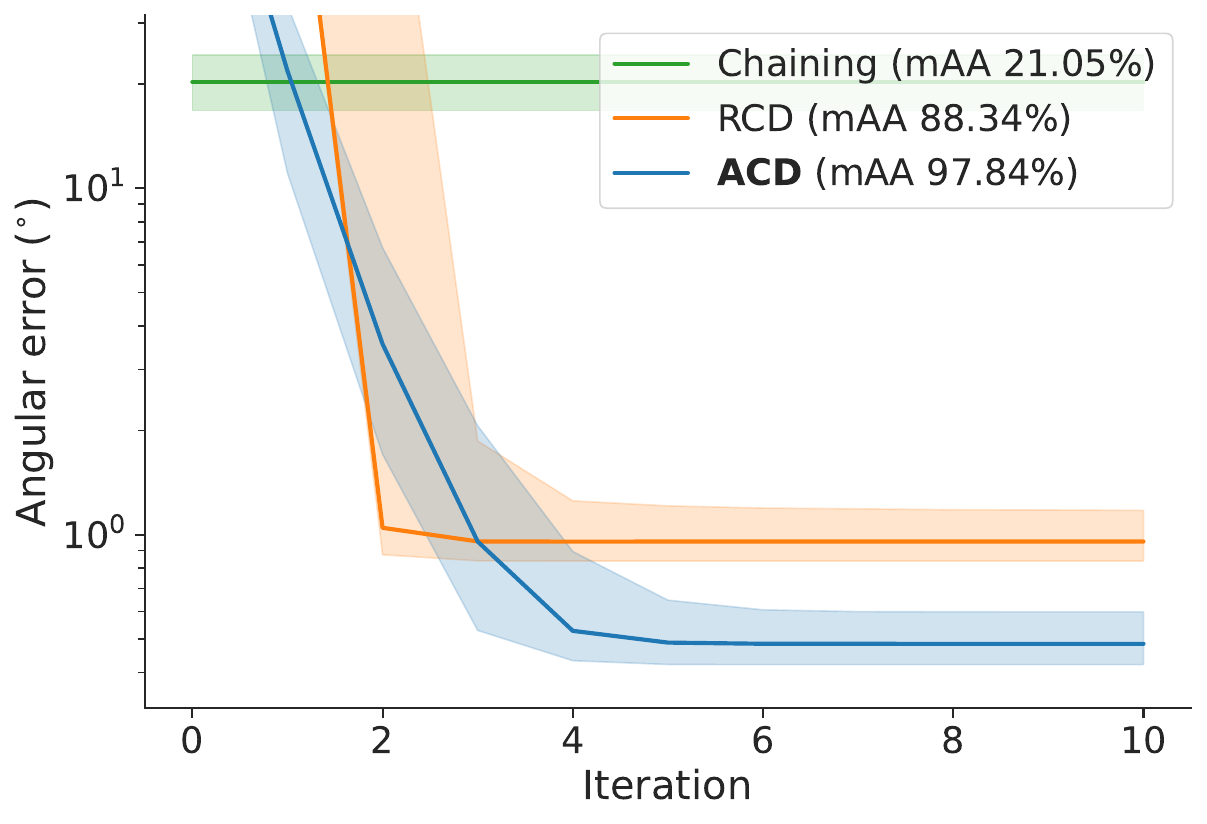}
\hspace{1em}
\includegraphics[width=0.42\textwidth]{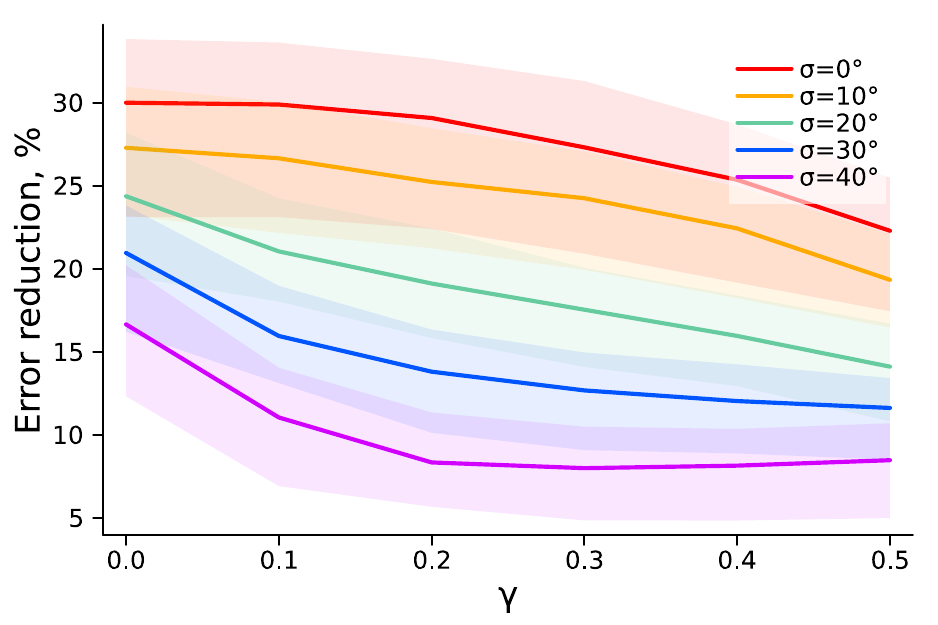}
\label{fig:synthetic}
\end{figure}

To generate synthetic scenes, we mainly follow the pipeline of~\cite{olsson2025} with the difference that the Hessian eigenvalues are sampled from $\mathcal U(a,b), ~b\sim \mathcal U(2a,100a), ~a\sim\mathcal U(10,100)$, and we also synthesize the ``loop'' scenes (Figure \ref{fig:synthetic}, left) in addition to the general random scenes (Figure \ref{fig:synthetic}, right).
For the ``loop'' scenes, we generate a set of 100 cameras evenly spread out on a circle and connect each camera with its two neighbors to form relative rotation estimates. In the noiseless setting, this problem has multiple good local minima. We found that all block coordinate descent methods converge to any of these if started from random rotations.
Initializing Algorithm~\ref{alg:method} with zeros is however sufficient to find a good minimum. In Figure \ref{fig:synthetic} (left), we present the convergence results for 1000 noisy ``loop'' scenes. We compare ACD to a simple rotation chaining as a baseline\footnote{MST followed by chaining is often used as an initialization strategy in robust rotation averaging \cite{chatterjee2013efficient,chatterjee2017,Zhang2023}.} and isotropic rotation averaging using RCD. We show median and interquartile range (IQR) of the angular error with respect to ground truth over the iterations. We also report the mean (over scenes) average (over cameras) accuracy, mAA. In a few iterations, ACD converges a more accurate solution.

\begin{table}[b]
\centering
\caption{Accuracy of non-robust rotation averaging on structure from motion datasets~\cite{olsson2011}.}
\resizebox{\textwidth}{!}{
\teenytiny
\begin{tabular}{rlllVlllVlll}
\toprule
 & \multicolumn{3}{c}{Angl. err. $\left( ^{\circ}\right)$ $\downarrow$} & \multicolumn{3}{c}{AUC @ 1$^{\circ}$ $\left(\%\right)$ $\uparrow$} & \multicolumn{3}{c}{AUC @ 5$^{\circ}$ $\left(\%\right)$ $\uparrow$} \\
\cmidrule(lr){2-4}\cmidrule(lr){5-7}\cmidrule(lr){8-10}
 & RCD & Shonan & \textbf{ACD} & RCD & Shonan & \textbf{ACD} & RCD & Shonan & \textbf{ACD} \\
 \midrule
LU Sphinx &   0.46 &   0.46 & \cellcolor{green1} \color{white}   0.36 &  97.14 &  97.14 & \cellcolor{green1} \color{white}  98.57 & 100.00 & 100.00 & 100.00 \\
Round Church &   0.59 &   0.59 & \cellcolor{green1} \color{white}   0.49 &  89.13 &  89.13 & \cellcolor{green1} \color{white}  94.78 & 100.00 & 100.00 & 100.00 \\
UWO &   1.19 &   1.19 & \cellcolor{green1} \color{white}   0.88 &  70.17 &  70.53 & \cellcolor{green1} \color{white}  87.72 &  99.12 &  99.12 & \cellcolor{green1} \color{white} 100.00 \\
Nikolai I &   0.48 &   0.48 & \cellcolor{green1} \color{white}   0.22 &  96.94 &  96.94 & \cellcolor{green1} \color{white} 100.00 & 100.00 & 100.00 & 100.00 \\
Vercingetorix &   1.54 &   1.56 & \cellcolor{green1} \color{white}   1.41 &  55.07 &  50.87 & \cellcolor{green1} \color{white}  60.29 &  98.55 &  98.55 &  98.55 \\
Eglise Du Dome &   0.24 &   0.24 & \cellcolor{green1} \color{white}   0.18 & 100.00 & 100.00 & 100.00 & 100.00 & 100.00 & 100.00 \\
King’s College &   0.78 &   0.77 & \cellcolor{green1} \color{white}   0.67 &  79.22 &  78.70 & \cellcolor{green1} \color{white}  82.86 & 100.00 & 100.00 & 100.00 \\
Kronan & \cellcolor{green1} \color{white}   0.76 & \cellcolor{green1} \color{white}   0.76 &   1.28 & \cellcolor{green1} \color{white}  94.66 & \cellcolor{green1} \color{white}  94.66 &  82.44 &  99.24 &  99.24 & \cellcolor{green1} \color{white} 100.00 \\
Alcatraz &   0.62 &   0.62 & \cellcolor{green1} \color{white}   0.52 &  90.23 &  90.23 & \cellcolor{green1} \color{white}  93.23 & 100.00 & 100.00 & 100.00 \\
Museum Barcelona &   0.79 &   0.79 & \cellcolor{green1} \color{white}   0.46 &  79.70 &  79.70 & \cellcolor{green1} \color{white}  99.25 & 100.00 & 100.00 & 100.00 \\
Temple Singapore &   0.85 &   0.86 & \cellcolor{green1} \color{white}   0.52 &  75.80 &  75.80 & \cellcolor{green1} \color{white}  95.67 & 100.00 & 100.00 & 100.00 \\
\midrule
\emph{Average} &   0.75 &   0.76 & \cellcolor{green1} \color{white}   0.64 &  84.37 &  83.97 & \cellcolor{green1} \color{white}  90.44 &  99.72 &  99.72 & \cellcolor{green1} \color{white}  99.87 \\
\bottomrule
\end{tabular}
}
\label{tab:RA}
\end{table}

\begin{table}[t]
\centering
\caption{Accuracy of non-robust rotation averaging on ETH3D MVS (DSLR)~\cite{schops2017multi}.}
\resizebox{\textwidth}{!}{
\teenytiny
\begin{tabular}{rlllVlllVlll}
\toprule
 & \multicolumn{3}{c}{Angl. err. $\left( ^{\circ}\right)$ $\downarrow$} & \multicolumn{3}{c}{AUC @ 1$^{\circ}$ $\left(\%\right)$ $\uparrow$} & \multicolumn{3}{c}{AUC @ 5$^{\circ}$ $\left(\%\right)$ $\uparrow$} \\
\cmidrule(lr){2-4}\cmidrule(lr){5-7}\cmidrule(lr){8-10}
 & RCD & Shonan & \textbf{ACD} & RCD & Shonan & \textbf{ACD} & RCD & Shonan & \textbf{ACD} \\
 \midrule
courtyard &   0.28 &   0.28 & \cellcolor{green1} \color{white}   0.18 &  97.37 &  97.37 & \cellcolor{green1} \color{white} 100.00 & 100.00 & 100.00 & 100.00 \\
delivery area &   0.46 &   0.51 & \cellcolor{green1} \color{white}   0.14 &  97.73 &  98.86 & \cellcolor{green1} \color{white} 100.00 & 100.00 & 100.00 & 100.00 \\
electro & \cellcolor{green1} \color{white}   0.33 & \cellcolor{green1} \color{white}   0.33 &   0.61 & \cellcolor{green1} \color{white}  97.78 & \cellcolor{green1} \color{white}  97.78 &  88.89 & 100.00 & 100.00 & 100.00 \\
facade &   0.19 &   0.19 & \cellcolor{green1} \color{white}   0.12 & 100.00 & 100.00 & 100.00 & 100.00 & 100.00 & 100.00 \\
meadow & \cellcolor{green1} \color{white}   0.96 & \cellcolor{green1} \color{white}   0.96 &   1.32 &  73.33 &  73.33 & \cellcolor{green1} \color{white}  80.00 & 100.00 & 100.00 & 100.00 \\
office &   1.24 &   1.24 & \cellcolor{green1} \color{white}   0.69 &  88.46 &  88.46 & \cellcolor{green1} \color{white}  92.31 &  96.15 &  96.15 & \cellcolor{green1} \color{white} 100.00 \\
pipes &   0.93 &   0.93 & \cellcolor{green1} \color{white}   0.66 &  85.71 &  85.71 & \cellcolor{green1} \color{white}  92.86 & 100.00 & 100.00 & 100.00 \\
playground &   0.11 &   0.11 & \cellcolor{green1} \color{white}   0.05 & 100.00 & 100.00 & 100.00 & 100.00 & 100.00 & 100.00 \\
relief &   0.38 &   0.38 & \cellcolor{green1} \color{white}   0.32 &  93.55 &  93.55 & \cellcolor{green1} \color{white}  96.77 & 100.00 & 100.00 & 100.00 \\
relief 2 &   0.44 &   0.44 & \cellcolor{green1} \color{white}   0.19 &  93.55 &  93.55 & \cellcolor{green1} \color{white} 100.00 & 100.00 & 100.00 & 100.00 \\
terrace &   0.79 &   0.79 & \cellcolor{green1} \color{white}   0.76 &  91.30 &  91.30 & \cellcolor{green1} \color{white}  95.65 & 100.00 & 100.00 & 100.00 \\
terrains &   0.39 &   0.38 & \cellcolor{green1} \color{white}   0.27 &  95.24 &  95.24 & \cellcolor{green1} \color{white} 100.00 & 100.00 & 100.00 & 100.00 \\
botanical garden &   0.93 &   1.09 & \cellcolor{green1} \color{white}   0.63 &  73.33 &  55.33 & \cellcolor{green1} \color{white}  96.67 & 100.00 & 100.00 & 100.00 \\
boulders &   0.27 &   0.27 & \cellcolor{green1} \color{white}   0.07 & 100.00 & 100.00 & 100.00 & 100.00 & 100.00 & 100.00 \\
bridge & \cellcolor{green1} \color{white}   0.16 & \cellcolor{green1} \color{white}   0.16 &   0.35 & \cellcolor{green1} \color{white} 100.00 & \cellcolor{green1} \color{white} 100.00 &  98.91 & 100.00 & 100.00 & 100.00 \\
door &   0.01 &   0.01 &   0.01 & 100.00 & 100.00 & 100.00 & 100.00 & 100.00 & 100.00 \\
exhibition hall &   0.19 &   0.19 & \cellcolor{green1} \color{white}   0.10 & 100.00 & 100.00 & 100.00 & 100.00 & 100.00 & 100.00 \\
lecture room &   0.23 &   0.23 & \cellcolor{green1} \color{white}   0.16 & 100.00 & 100.00 & 100.00 & 100.00 & 100.00 & 100.00 \\
living room & \cellcolor{green1} \color{white}   0.83 & \cellcolor{green1} \color{white}   0.83 &   0.94 &  84.61 &  84.61 & \cellcolor{green1} \color{white}  87.69 & 100.00 & 100.00 & 100.00 \\
observatory &   0.05 &   0.05 & \cellcolor{green1} \color{white}   0.04 & 100.00 & 100.00 & 100.00 & 100.00 & 100.00 & 100.00 \\
old computer &   0.22 &   0.22 & \cellcolor{green1} \color{white}   0.19 & 100.00 & 100.00 & 100.00 & 100.00 & 100.00 & 100.00 \\
statue &   0.07 &   0.07 & \cellcolor{green1} \color{white}   0.01 & 100.00 & 100.00 & 100.00 & 100.00 & 100.00 & 100.00 \\
terrace 2 &   0.01 &   0.01 &   0.01 & 100.00 & 100.00 & 100.00 & 100.00 & 100.00 & 100.00 \\
\midrule
\emph{Average} &   0.41 &   0.42 & \cellcolor{green1} \color{white}   0.34 &  94.43 &  93.70 & \cellcolor{green1} \color{white}  96.95 &  99.83 &  99.83 & \cellcolor{green1} \color{white} 100.00 \\
\bottomrule
\end{tabular}
}
\label{tab:ETH3D}
\end{table}

For the general scenes, we randomly sample absolute orientations of 100 cameras and a fraction $p \sim \mathcal U(0.1,1)$ of observed relative rotations.
In Figure \ref{fig:synthetic} (right), we studied how the Hessians' eigenvector perturbations---rotating around a random axis by $\mathcal N(0,\sigma)$ degrees---and eigenvalue perturbations---\ie adding uniform noise $\mathcal U(0,\gamma\cdot\text{avg.\ EV})$---affect the performance. We show median and IQR of the error reduction in $\%$ compared to the isotropic RCD method. We see that small perturbations slightly affect the accuracy, however the error reduction is still quite significant (from around $30\%$ down to around $10\%$).

\begin{table}[b]
\centering
\caption{Accuracy of robust rotation averaging on structure from motion datasets~\cite{olsson2011}.}
\resizebox{\textwidth}{!}{
\teenytiny
\begin{tabular}{rllllVllllVllll}
\toprule
 & \multicolumn{4}{c}{Angl. err. $\left( ^{\circ}\right)$ $\downarrow$} & \multicolumn{4}{c}{AUC @ 1$^{\circ}$ $\left(\%\right)$ $\uparrow$} & \multicolumn{4}{c}{AUC @ 5$^{\circ}$ $\left(\%\right)$ $\uparrow$} \\
\cmidrule(lr){2-5}\cmidrule(lr){6-9}\cmidrule(lr){10-13}
& \ttiny L1IRLS & \ttiny Zhang & \ttiny \textbf{ACD}\textsubscript{\ttiny IRLS} & \ttiny \textbf{ACD}\textsubscript{\ttiny AIRLS} & \ttiny L1IRLS & \ttiny Zhang & \ttiny \textbf{ACD}\textsubscript{\ttiny IRLS} & \ttiny \textbf{ACD}\textsubscript{\ttiny AIRLS} & \ttiny L1IRLS & \ttiny Zhang & \ttiny \textbf{ACD}\textsubscript{\ttiny IRLS} & \ttiny \textbf{ACD}\textsubscript{\ttiny AIRLS} \\
 \midrule
LU Sphinx &   0.41 & \cellcolor{green1} \color{white}   0.37 &   0.41 & \cellcolor{green1} \color{white}   0.37 & \cellcolor{green1} \color{white}  98.57 & \cellcolor{green1} \color{white}  98.57 & \cellcolor{green1} \color{white}  98.57 &  97.14 & 100.00 & 100.00 & 100.00 & 100.00 \\
Round Church &   0.59 & \cellcolor{green1} \color{white}   0.53 &   0.59 &   0.55 &  92.39 &  93.48 &  92.39 & \cellcolor{green1} \color{white}  94.56 & 100.00 & 100.00 & 100.00 & 100.00 \\
UWO &   1.08 &  13.52 &   1.08 & \cellcolor{green1} \color{white}   0.55 &  41.23 &  52.63 &  41.23 & \cellcolor{green1} \color{white}  99.12 & \cellcolor{green1} \color{white} 100.00 &  95.61 & \cellcolor{green1} \color{white} 100.00 & \cellcolor{green1} \color{white} 100.00 \\
Nikolai I &   0.33 &   5.17 &   0.33 & \cellcolor{green1} \color{white}   0.27 & \cellcolor{green1} \color{white} 100.00 &  98.98 & \cellcolor{green1} \color{white} 100.00 & \cellcolor{green1} \color{white} 100.00 & \cellcolor{green1} \color{white} 100.00 &  98.98 & \cellcolor{green1} \color{white} 100.00 & \cellcolor{green1} \color{white} 100.00 \\
Vercingetorix & \cellcolor{green1} \color{white}   0.47 &  38.20 & \cellcolor{green1} \color{white}   0.47 &   0.54 & \cellcolor{green1} \color{white}  94.20 &   0.00 & \cellcolor{green1} \color{white}  94.20 &  92.75 & \cellcolor{green1} \color{white} 100.00 &   0.00 & \cellcolor{green1} \color{white} 100.00 & \cellcolor{green1} \color{white} 100.00 \\
Eglise Du Dome &   0.22 &   0.18 &   0.22 & \cellcolor{green1} \color{white}   0.17 & 100.00 & 100.00 & 100.00 & 100.00 & 100.00 & 100.00 & 100.00 & 100.00 \\
King’s College &   0.79 & \cellcolor{green1} \color{white}   0.70 &   0.79 &   0.73 &  76.62 & \cellcolor{green1} \color{white}  81.82 &  76.62 &  80.52 & 100.00 & 100.00 & 100.00 & 100.00 \\
Kronan &   0.40 &   7.44 &   0.40 & \cellcolor{green1} \color{white}   0.35 &  97.71 &  87.02 &  97.71 & \cellcolor{green1} \color{white} 100.00 & \cellcolor{green1} \color{white} 100.00 &  98.47 & \cellcolor{green1} \color{white} 100.00 & \cellcolor{green1} \color{white} 100.00 \\
Alcatraz &   0.58 & \cellcolor{green1} \color{white}   0.49 &   0.58 &   0.53 &  91.73 & \cellcolor{green1} \color{white}  93.23 &  91.73 &  91.73 & 100.00 & 100.00 & 100.00 & 100.00 \\
Museum Barcelona & \cellcolor{green1} \color{white}   0.43 &   8.29 & \cellcolor{green1} \color{white}   0.43 & \cellcolor{green1} \color{white}   0.43 & \cellcolor{green1} \color{white} 100.00 &  99.25 & \cellcolor{green1} \color{white} 100.00 &  97.74 & \cellcolor{green1} \color{white} 100.00 &  99.25 & \cellcolor{green1} \color{white} 100.00 & \cellcolor{green1} \color{white} 100.00 \\
Temple Singapore &   0.59 &   0.57 &   0.59 & \cellcolor{green1} \color{white}   0.56 &  91.72 &  92.99 &  91.72 & \cellcolor{green1} \color{white}  96.18 & 100.00 & 100.00 & 100.00 & 100.00 \\
\midrule
\emph{Average} &   0.53 &   6.86 &   0.53 & \cellcolor{green1} \color{white}   0.46 &  89.47 &  81.63 &  89.47 & \cellcolor{green1} \color{white}  95.43 & \cellcolor{green1} \color{white} 100.00 &  90.21 & \cellcolor{green1} \color{white} 100.00 & \cellcolor{green1} \color{white} 100.00 \\
\bottomrule
\end{tabular}
}
\label{tab:RA_robust}
\end{table}

\begin{table}[t]
\centering
\caption{Accuracy of robust rotation averaging on ETH3D MVS (DSLR)~\cite{schops2017multi}.}
\resizebox{\textwidth}{!}{
\teenytiny
\begin{tabular}{rllllVllllVllll}
\toprule
 & \multicolumn{4}{c}{Angl. err. $\left( ^{\circ}\right)$ $\downarrow$} & \multicolumn{4}{c}{AUC @ 1$^{\circ}$ $\left(\%\right)$ $\uparrow$} & \multicolumn{4}{c}{AUC @ 5$^{\circ}$ $\left(\%\right)$ $\uparrow$} \\
\cmidrule(lr){2-5}\cmidrule(lr){6-9}\cmidrule(lr){10-13}
 & \ttiny L1IRLS & \ttiny Zhang & \ttiny \textbf{ACD}\textsubscript{\ttiny IRLS} & \ttiny \textbf{ACD}\textsubscript{\ttiny AIRLS} & \ttiny L1IRLS & \ttiny Zhang & \ttiny \textbf{ACD}\textsubscript{\ttiny IRLS} & \ttiny \textbf{ACD}\textsubscript{\ttiny AIRLS} & \ttiny L1IRLS & \ttiny Zhang & \ttiny \textbf{ACD}\textsubscript{\ttiny IRLS} & \ttiny \textbf{ACD}\textsubscript{\ttiny AIRLS} \\
 \midrule
courtyard &   0.17 &   0.17 &   0.17 & \cellcolor{green1} \color{white}   0.15 & 100.00 & 100.00 & 100.00 & 100.00 & 100.00 & 100.00 & 100.00 & 100.00 \\
delivery area &   0.11 &   0.10 &   0.11 & \cellcolor{green1} \color{white}   0.09 & 100.00 & 100.00 & 100.00 & 100.00 & 100.00 & 100.00 & 100.00 & 100.00 \\
electro & \cellcolor{green1} \color{white}   0.20 &   0.40 & \cellcolor{green1} \color{white}   0.20 &   0.77 & \cellcolor{green1} \color{white} 100.00 &  97.78 & \cellcolor{green1} \color{white} 100.00 &  93.33 & 100.00 & 100.00 & 100.00 & 100.00 \\
facade &   0.14 & \cellcolor{green1} \color{white}   0.11 &   0.14 &   0.12 & 100.00 & 100.00 & 100.00 & 100.00 & 100.00 & 100.00 & 100.00 & 100.00 \\
meadow & \cellcolor{green1} \color{white}   0.47 &   0.70 & \cellcolor{green1} \color{white}   0.47 &   0.68 &  86.67 & \cellcolor{green1} \color{white}  93.33 &  86.67 &  86.67 & 100.00 & 100.00 & 100.00 & 100.00 \\
office & \cellcolor{green1} \color{white}   0.26 &   4.50 & \cellcolor{green1} \color{white}   0.26 &   0.73 & \cellcolor{green1} \color{white} 100.00 &  69.23 & \cellcolor{green1} \color{white} 100.00 &  96.15 & \cellcolor{green1} \color{white} 100.00 &  96.15 & \cellcolor{green1} \color{white} 100.00 & \cellcolor{green1} \color{white} 100.00 \\
pipes & \cellcolor{green1} \color{white}   0.18 &   5.13 & \cellcolor{green1} \color{white}   0.18 &   0.69 & \cellcolor{green1} \color{white} 100.00 &   0.00 & \cellcolor{green1} \color{white} 100.00 &  92.86 & \cellcolor{green1} \color{white} 100.00 &  92.86 & \cellcolor{green1} \color{white} 100.00 & \cellcolor{green1} \color{white} 100.00 \\
playground &   0.08 &   0.05 &   0.08 & \cellcolor{green1} \color{white}   0.04 & 100.00 & 100.00 & 100.00 & 100.00 & 100.00 & 100.00 & 100.00 & 100.00 \\
relief & \cellcolor{green1} \color{white}   0.07 &   0.16 & \cellcolor{green1} \color{white}   0.07 &   0.14 & 100.00 & 100.00 & 100.00 & 100.00 & 100.00 & 100.00 & 100.00 & 100.00 \\
relief 2 & \cellcolor{green1} \color{white}   0.07 &   3.36 & \cellcolor{green1} \color{white}   0.07 &   0.15 & \cellcolor{green1} \color{white} 100.00 &  96.77 & \cellcolor{green1} \color{white} 100.00 & \cellcolor{green1} \color{white} 100.00 & \cellcolor{green1} \color{white} 100.00 &  96.77 & \cellcolor{green1} \color{white} 100.00 & \cellcolor{green1} \color{white} 100.00 \\
terrace &   0.12 &   0.43 &   0.12 & \cellcolor{green1} \color{white}   0.07 & \cellcolor{green1} \color{white} 100.00 &  95.65 & \cellcolor{green1} \color{white} 100.00 & \cellcolor{green1} \color{white} 100.00 & 100.00 & 100.00 & 100.00 & 100.00 \\
terrains &   0.13 & \cellcolor{green1} \color{white}   0.10 &   0.13 &   0.26 & 100.00 & 100.00 & 100.00 & 100.00 & 100.00 & 100.00 & 100.00 & 100.00 \\
botanical garden &   0.84 &   0.61 &   0.84 & \cellcolor{green1} \color{white}   0.54 &  90.00 &  96.67 &  90.00 & \cellcolor{green1} \color{white} 100.00 & 100.00 & 100.00 & 100.00 & 100.00 \\
boulders &   0.11 &   0.06 &   0.11 & \cellcolor{green1} \color{white}   0.03 & 100.00 & 100.00 & 100.00 & 100.00 & 100.00 & 100.00 & 100.00 & 100.00 \\
bridge & \cellcolor{green1} \color{white}   0.09 &   0.13 & \cellcolor{green1} \color{white}   0.09 &   0.10 & 100.00 & 100.00 & 100.00 & 100.00 & 100.00 & 100.00 & 100.00 & 100.00 \\
door & \cellcolor{green1} \color{white}   0.01 &   0.02 & \cellcolor{green1} \color{white}   0.01 & \cellcolor{green1} \color{white}   0.01 & 100.00 & 100.00 & 100.00 & 100.00 & 100.00 & 100.00 & 100.00 & 100.00 \\
exhibition hall &   0.07 &   0.07 &   0.07 & \cellcolor{green1} \color{white}   0.06 & 100.00 & 100.00 & 100.00 & 100.00 & 100.00 & 100.00 & 100.00 & 100.00 \\
lecture room &   0.16 & \cellcolor{green1} \color{white}   0.10 &   0.16 &   0.16 & 100.00 & 100.00 & 100.00 & 100.00 & 100.00 & 100.00 & 100.00 & 100.00 \\
living room & \cellcolor{green1} \color{white}   0.15 &   4.59 & \cellcolor{green1} \color{white}   0.15 &   0.40 & \cellcolor{green1} \color{white} 100.00 &  83.08 & \cellcolor{green1} \color{white} 100.00 & \cellcolor{green1} \color{white} 100.00 & \cellcolor{green1} \color{white} 100.00 &  95.39 & \cellcolor{green1} \color{white} 100.00 & \cellcolor{green1} \color{white} 100.00 \\
observatory &   0.05 & \cellcolor{green1} \color{white}   0.04 &   0.05 & \cellcolor{green1} \color{white}   0.04 & 100.00 & 100.00 & 100.00 & 100.00 & 100.00 & 100.00 & 100.00 & 100.00 \\
old computer &   0.10 &   0.11 &   0.10 & \cellcolor{green1} \color{white}   0.09 & 100.00 & 100.00 & 100.00 & 100.00 & 100.00 & 100.00 & 100.00 & 100.00 \\
statue &   0.07 & \cellcolor{green1} \color{white}   0.01 &   0.07 &   0.02 & 100.00 & 100.00 & 100.00 & 100.00 & 100.00 & 100.00 & 100.00 & 100.00 \\
terrace 2 &   0.01 &   0.01 &   0.01 &   0.01 & 100.00 & 100.00 & 100.00 & 100.00 & 100.00 & 100.00 & 100.00 & 100.00 \\
\midrule
\emph{Average} & \cellcolor{green1} \color{white}   0.16 &   0.91 & \cellcolor{green1} \color{white}   0.16 &   0.23 & \cellcolor{green1} \color{white}  98.99 &  92.72 & \cellcolor{green1} \color{white}  98.99 &  98.65 & \cellcolor{green1} \color{white} 100.00 &  99.18 & \cellcolor{green1} \color{white} 100.00 & \cellcolor{green1} \color{white} 100.00 \\
\bottomrule
\end{tabular}
}
\label{tab:ETH3D_robust}
\end{table}

\subsection{Experiments on real data}
We compare the proposed ACD to the following methods: (i) two SDP solvers optimizing isotropic and anisotropic chordal distances --- O(3)~\cite{eriksson2018} and cSO(3)~\cite{olsson2025}, respectively; (ii) two dedicated solvers optimizing isotropic chordal distances --- RCD~\cite{parra2021} and Shonan~\cite{dellaert2020}, (iii) a local robust rotation averaging method of Chatterjee 
\etal~\cite{chatterjee2013efficient} optimizing isotropic distances in the axis-angle representation space using L1 averaging followed by iteratively reweighted least squares (L1IRLS), and (iv) a local robust anisotropic rotation averaging method of Zhang~\etal~\cite{Zhang2023} anisotropic distances in the axis-angle representation space using MAGSAC (Zhang). We use the author's implementations of their methods, where RCD, L1IRLS, and Zhang are initialized using MST, while Shonan, O(3), and cSO(3) start from random initialization. Our method is initialized with zeros. Since ACD employs camera shuffling (see Algorithm~\ref{alg:method}), we run it 10 times on each dataset.

We evaluate rotation averaging methods on structure from motion (SfM) datasets~\cite{olsson2011} and ETH3D MVS (DSLR) datasets~\cite{schops2017multi}. 
We obtain pairwise relative rotations and Hessian estimates via
matched SIFT points for the SfM datasets and data provided as part of the ETH3D datasets. 
We then follow a standard two-view estimation pipeline, where we (i) obtain the initial relative pose with RANSAC and the 5 point solver~\cite{nister2004}, (ii) refine the relative pose together with the 3D points using two-view bundle adjustment~\cite{triggs1999}, i.e. by minimizing the sum-of-squared reprojection errors, and (iii) use the Jacobian $J$ of the residual function to obtain the full Hessian and marginalize out the 3D points and relative position via Schur complement~\cite{olsson2011}. For a fair comparison (\ie to alleviate potential differences due to different two-view costs) we use the same Hessians in both ACD and Zhang~\etal~\cite{Zhang2023}.

Tables \ref{tab:RA} and \ref{tab:ETH3D} compare ACD to the other non-robust rotation averaging solvers, RCD~\cite{parra2021} and Shonan~\cite{dellaert2020}. These methods are specifically designed to optimize isotropic chordal distances. Results for the SfM datasets are shown in Table \ref{tab:RA}, where the proposed method outperforms RCD and Shonan (that are both producing very similar results) in all but one dataset. These results are in line with those obtained for the corresponding SDP solvers in~\cite{olsson2025}. Note that there is a large improvement in the proportion of highly accurate solutions (\ie giving less than $1$ degree error).
Table \ref{tab:ETH3D} presents results for the ETH3D datasets, where ACD gives better estimates in the majority of the scenes. Similarly, we see a particular increase in the percentage of highly accurate solutions. Overall, the tendency is that anisotropic rotation averaging gives more accurate solutions (see also Section A in the Supplementary).

\subsection{Robust rotation averaging}
We test the proposed ACD solver as an initialization strategy for robust refinement and obtain two algorithms, ACD\textsubscript{\small IRLS} and ACD\textsubscript{\small AIRLS}.
ACD\textsubscript{\small IRLS} refers to applying iteratively reweighted least squares (see Algorithm 2 in~\cite{chatterjee2013efficient}).
The robust estimation stage of ACD\textsubscript{\small AIRLS} incorporates the uncertainties into the objective (similarly to~\cite{Zhang2023}).
Let $\omega(\cdot)$ extract the exponential map parameters, $\{\Delta \hat\omega_k\}$ encode the incremental updates of rotations $\{\mR_k\}$, and $\rho(x) = \frac{x^2}{x^2 + \tau^2}$ be the Geman-McClure kernel function.
The following objective is iteratively optimized by ACD\textsubscript{\small AIRLS}
\begin{align}
\min\nolimits_{\{\Delta \hat\omega_k\}} \sum\nolimits_{(i,j) \in \cal{E}} \rho\left(\|(\Delta\hat\omega_j - \Delta\hat\omega_i) - \omega(\mR_{j}\tr\tilde\mR_{ij}\mR_{i})\|_{\mH_{ij}}\right),
\label{eq:anisotrpic_IRLS}
\end{align}
followed by updating $\mR_k \leftarrow \mR_k e^{[\Delta\hat\omega_k]_\times}$.
In Tables \ref{tab:RA_robust} and \ref{tab:ETH3D_robust}, we compare these methods to the other robust rotation averaging approaches on SfM and ETH3D datasets, respectively. We see that the solution accuracy of ACD\textsubscript{\small IRLS} is identical to L1IRLS~\cite{chatterjee2013efficient}, and is higher than that of Zhang~\etal~\cite{Zhang2023}. ACD\textsubscript{\small AIRLS} outperforms the other methods on the majority of the SfM datasets and many ETH3D datasets. The method of Zhang~\etal~\cite{Zhang2023} gives poor solutions in a several cases, \eg UWO, Vercingetorix. We think that might be related to MAGSAC's hyperparameter that may need to be tuned for each dataset. On the other hand, ACD does not require hyperparameter tuning, and for the refinement step we used the same threshold parameter $\tau = 5^\circ$ of the robust kernel as in the author's implementation of~\cite{chatterjee2013efficient}.

\subsection{Runtime}
\begin{figure}[!b]
\centering
\caption{Runtimes of rotation averaging algorithms.}
\includegraphics[width=0.42\textwidth]{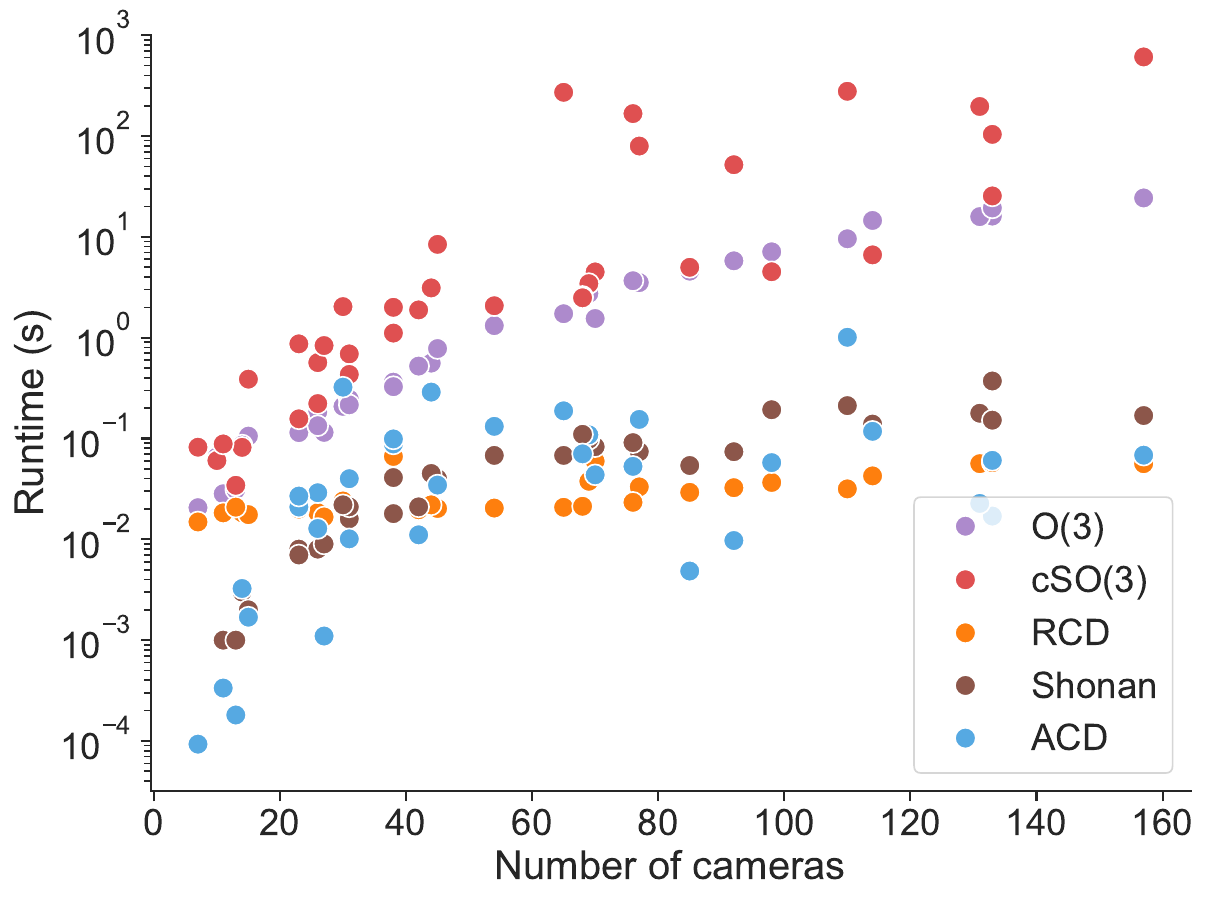}
\includegraphics[width=0.42\textwidth]{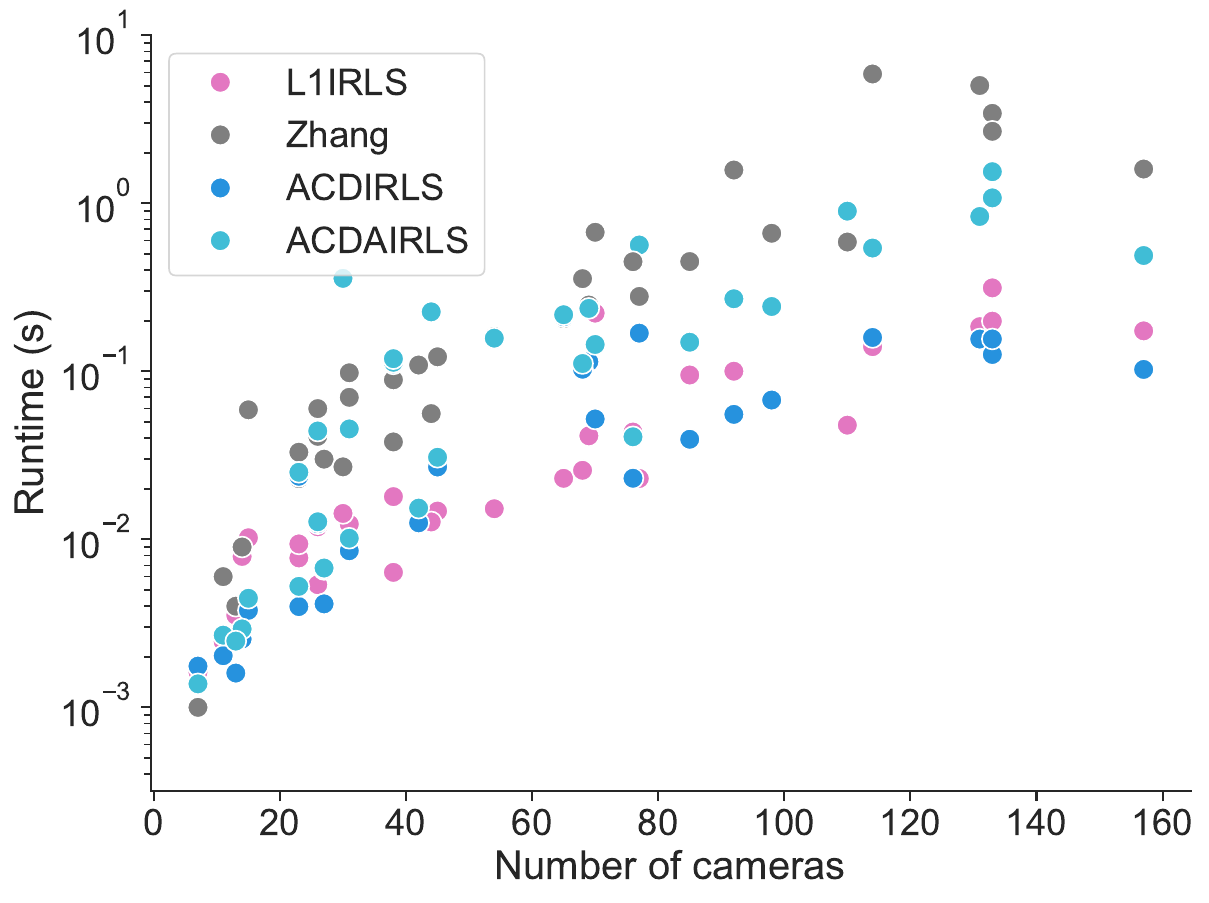}
\label{fig:runtimes}
\end{figure}

We further present the runtime of rotation averaging algorithms. Both absolute and relative tolerance in ACD are set to $10^{-12}$. As indicated in Figure~\ref{fig:runtimes}~(left), ACD is as fast and sometimes faster than the state-of-the-art non-robust solvers RCD~\cite{parra2021} and Shonan~\cite{dellaert2020}. More importantly, ACD is---not surprisingly---orders of magnitude faster than the anisotropic cSO(3)~\cite{olsson2025}. Further, Figure~\ref{fig:runtimes}~(right) shows that ACD\textsubscript{\small IRLS} is on par with the state-of-the-art robust method L1IRLS~\cite{chatterjee2013efficient} in terms of runtime, while ACD\textsubscript{\small AIRLS} has comparable runtimes with the method of Zhang~\etal~\cite{Zhang2023}.

\subsection{Ablation studies}\label{sec:ablation}
Table \ref{tab:ablation_study} compares different initializations of ACD: random rotation matrices (rand), all identities (id), all zeros (0), MST + chaining (mst), and presents corresponding results for the SDP solver of~\cite{olsson2025} (cSO(3)). We found that in practice all initializations performed similarly well on real data. Results for two datasets---King's College and facade---indicate that possibly ACD converged to a different solution than cSO(3). Those are however reasonably good judging by the obtained accuracies. Overall, in all runs, ACD gave accurate solutions.

\begin{table}[h]
\centering
\caption{Comparison to the SDP solver of \cite{olsson2025} (cSO(3)) and effects of different initializations on \textbf{ACD} solver: random rotations (rand), all identities (id), all zeros (0), MST+chaining (mst)}
\resizebox{\textwidth}{!}{
\teenytiny
\begin{tabular}{rlllllVlllll}
\toprule
 & \multicolumn{5}{c}{Angl. err. $\left( ^{\circ}\right)$ $\downarrow$} & \multicolumn{5}{c}{AA $\left(\%\right)$ $\uparrow$} \\
\cmidrule(lr){2-6}\cmidrule(lr){7-11}
 & cSO(3) & \textbf{ACD}–rand & \textbf{ACD}-id & \textbf{ACD}-0 & \textbf{ACD}-mst & cSO(3) & \textbf{ACD}–rand & \textbf{ACD}-id & \textbf{ACD}-0 & \textbf{ACD}-mst \\
 \midrule
\hspace*{-0.7cm}\multirow{11}{*}{\rotatebox{90}{SfM}}\hspace*{0.7cm} LU Sphinx  &   0.36 &   0.36 &   0.36 &   0.36 &   0.36 & \cellcolor{green1} \color{white}  98.76 &  98.75 & \cellcolor{green1} \color{white}  98.76 & \cellcolor{green1} \color{white}  98.76 &  98.73 \\
Round Church &   0.54 & \cellcolor{green1} \color{white}   0.48 & \cellcolor{green1} \color{white}   0.48 &   0.49 & \cellcolor{green1} \color{white}   0.48 & \cellcolor{green1} \color{white}  98.70 &  98.50 &  98.53 &  98.42 &  98.52 \\
UWO & \cellcolor{green1} \color{white}   0.86 &   0.89 &   0.89 &   0.88 & \cellcolor{green1} \color{white}   0.86 &  97.22 &  97.04 &  97.03 &  97.08 & \cellcolor{green1} \color{white}  97.24 \\
Nikolai I & \cellcolor{green1} \color{white}   0.22 & \cellcolor{green1} \color{white}   0.22 &   0.23 & \cellcolor{green1} \color{white}   0.22 & \cellcolor{green1} \color{white}   0.22 &  99.31 &  99.32 &  99.28 &  99.31 & \cellcolor{green1} \color{white}  99.34 \\
Vercingetorix &   1.42 &   1.41 & \cellcolor{green1} \color{white}   1.40 &   1.41 &   1.41 &  94.46 &  94.54 & \cellcolor{green1} \color{white}  94.57 &  94.55 & \cellcolor{green1} \color{white}  94.57 \\
Eglise Du Dome &   0.21 & \cellcolor{green1} \color{white}   0.18 & \cellcolor{green1} \color{white}   0.18 & \cellcolor{green1} \color{white}   0.18 & \cellcolor{green1} \color{white}   0.18 &  99.39 &  99.52 &  99.51 &  99.52 & \cellcolor{green1} \color{white}  99.54 \\
King’s College & \cellcolor{green1} \color{white}   0.37 &   0.68 &   0.67 &   0.67 &   0.68 & \cellcolor{green1} \color{white}  98.69 &  97.18 &  97.27 &  97.24 &  97.19 \\
Kronan &   1.38 & \cellcolor{green1} \color{white}   1.25 &   1.26 &   1.28 &   1.27 &  96.10 & \cellcolor{green1} \color{white}  96.46 &  96.44 &  96.38 &  96.41 \\
Alcatraz & \cellcolor{green1} \color{white}   0.45 &   0.52 &   0.52 &   0.52 &   0.52 & \cellcolor{green1} \color{white}  98.50 &  98.27 &  98.27 &  98.28 &  98.29 \\
Museum Barcelona &   0.46 &   0.46 & \cellcolor{green1} \color{white}   0.45 &   0.46 &   0.46 &  98.81 &  98.83 & \cellcolor{green1} \color{white}  98.89 &  98.81 &  98.74 \\
Temple Singapore &   0.55 &   0.55 & \cellcolor{green1} \color{white}   0.52 & \cellcolor{green1} \color{white}   0.52 &   0.59 & \cellcolor{green1} \color{white}  98.34 &  97.99 &  98.16 &  98.14 &  97.83 \\
\midrule
\hspace*{-0.7cm}\multirow{21}{*}{\rotatebox{90}{ETH3D}}\hspace*{0.7cm} courtyard &   0.18 &   0.18 &   0.18 &   0.18 &   0.18 & \cellcolor{green1} \color{white}  99.54 &  99.52 &  99.53 &  99.52 &  99.53 \\
delivery area &   0.13 &   0.16 &   0.16 &   0.14 & \cellcolor{green1} \color{white}   0.12 & \cellcolor{green1} \color{white}  99.77 &  99.53 &  99.52 &  99.62 &  99.76 \\
electro &   0.71 & \cellcolor{green1} \color{white}   0.61 & \cellcolor{green1} \color{white}   0.61 & \cellcolor{green1} \color{white}   0.61 & \cellcolor{green1} \color{white}   0.61 &  98.21 &  98.45 & \cellcolor{green1} \color{white}  98.46 &  98.44 &  98.44 \\
facade &   8.31 & \cellcolor{green1} \color{white}   0.12 &   0.13 & \cellcolor{green1} \color{white}   0.12 & \cellcolor{green1} \color{white}   0.12 &  95.10 &  99.80 & \cellcolor{green1} \color{white}  99.81 & \cellcolor{green1} \color{white}  99.81 & \cellcolor{green1} \color{white}  99.81 \\
meadow &   1.43 & \cellcolor{green1} \color{white}   1.32 & \cellcolor{green1} \color{white}   1.32 & \cellcolor{green1} \color{white}   1.32 & \cellcolor{green1} \color{white}   1.32 &  95.10 &  95.45 &  95.44 &  95.46 & \cellcolor{green1} \color{white}  95.47 \\
office &   0.69 &   0.69 &   0.69 &   0.69 &   0.69 & \cellcolor{green1} \color{white}  98.08 & \cellcolor{green1} \color{white}  98.08 & \cellcolor{green1} \color{white}  98.08 & \cellcolor{green1} \color{white}  98.08 &  98.06 \\
pipes &   0.66 &   0.66 &   0.66 &   0.66 &   0.66 &  97.79 &  97.79 &  97.79 &  97.79 &  97.79 \\
playground &   0.05 &   0.05 &   0.05 &   0.05 &   0.05 & \cellcolor{green1} \color{white}  99.99 &  99.98 & \cellcolor{green1} \color{white}  99.99 &  99.97 & \cellcolor{green1} \color{white}  99.99 \\
relief &   0.33 & \cellcolor{green1} \color{white}   0.32 & \cellcolor{green1} \color{white}   0.32 & \cellcolor{green1} \color{white}   0.32 & \cellcolor{green1} \color{white}   0.32 &  99.26 &  99.42 & \cellcolor{green1} \color{white}  99.43 & \cellcolor{green1} \color{white}  99.43 & \cellcolor{green1} \color{white}  99.43 \\
relief 2 &   0.20 & \cellcolor{green1} \color{white}   0.19 & \cellcolor{green1} \color{white}   0.19 & \cellcolor{green1} \color{white}   0.19 & \cellcolor{green1} \color{white}   0.19 &  99.48 &  99.54 &  99.53 &  99.54 & \cellcolor{green1} \color{white}  99.58 \\
terrace & \cellcolor{green1} \color{white}   0.73 &   0.76 &   0.76 &   0.76 &   0.76 & \cellcolor{green1} \color{white}  98.63 &  98.54 &  98.54 &  98.53 &  98.54 \\
terrains &   0.34 & \cellcolor{green1} \color{white}   0.27 & \cellcolor{green1} \color{white}   0.27 & \cellcolor{green1} \color{white}   0.27 & \cellcolor{green1} \color{white}   0.27 &  98.80 &  99.19 & \cellcolor{green1} \color{white}  99.21 &  99.20 &  99.20 \\
botanical garden &   0.58 &   0.58 &   0.58 &   0.63 & \cellcolor{green1} \color{white}   0.56 &  97.48 &  97.48 &  97.49 &  97.22 & \cellcolor{green1} \color{white}  97.55 \\
boulders &   0.09 & \cellcolor{green1} \color{white}   0.06 &   0.07 &   0.07 &   0.07 &  99.92 & \cellcolor{green1} \color{white}  99.97 &  99.94 &  99.96 &  99.95 \\
bridge & \cellcolor{green1} \color{white}   0.32 &   0.38 &   0.33 &   0.35 &   0.33 & \cellcolor{green1} \color{white}  98.82 &  98.49 &  98.78 &  98.66 &  98.74 \\
door &   0.02 & \cellcolor{green1} \color{white}   0.01 &   0.02 & \cellcolor{green1} \color{white}   0.01 & \cellcolor{green1} \color{white}   0.01 & 100.00 & 100.00 & 100.00 & 100.00 & 100.00 \\
exhibition hall &   0.10 &   0.10 &   0.10 &   0.10 &   0.10 &  99.82 & \cellcolor{green1} \color{white}  99.83 & \cellcolor{green1} \color{white}  99.83 & \cellcolor{green1} \color{white}  99.83 & \cellcolor{green1} \color{white}  99.83 \\
lecture room &   0.16 &   0.16 &   0.16 &   0.16 &   0.16 & \cellcolor{green1} \color{white}  99.72 &  99.71 &  99.71 &  99.71 & \cellcolor{green1} \color{white}  99.72 \\
living room &   1.22 &   1.09 & \cellcolor{green1} \color{white}   0.92 &   0.94 &   0.94 &  96.16 &  96.58 & \cellcolor{green1} \color{white}  97.12 &  97.06 &  97.07 \\
observatory &   0.07 & \cellcolor{green1} \color{white}   0.04 &   0.05 & \cellcolor{green1} \color{white}   0.04 & \cellcolor{green1} \color{white}   0.04 &  99.96 & \cellcolor{green1} \color{white} 100.00 & \cellcolor{green1} \color{white} 100.00 & \cellcolor{green1} \color{white} 100.00 & \cellcolor{green1} \color{white} 100.00 \\
old computer &   0.22 & \cellcolor{green1} \color{white}   0.18 &   0.20 &   0.19 &   0.20 &  99.31 & \cellcolor{green1} \color{white}  99.57 &  99.35 &  99.48 &  99.37 \\
statue &   0.01 &   0.01 &   0.01 &   0.01 &   0.01 & 100.00 & 100.00 & 100.00 & 100.00 & 100.00 \\
terrace 2 &   0.01 &   0.01 &   0.01 &   0.01 &   0.01 & 100.00 & 100.00 & 100.00 & 100.00 & 100.00 \\
\midrule
\emph{Average} &   0.69 &   0.44 & \cellcolor{green1} \color{white}   0.43 &   0.44 & \cellcolor{green1} \color{white}   0.43 &  98.51 &  98.63 & \cellcolor{green1} \color{white}  98.65 &  98.64 & \cellcolor{green1} \color{white}  98.65 \\
\bottomrule
\end{tabular}
}
\label{tab:ablation_study}
\end{table}

\section{Conclusions}
In this paper, we address the problem of fast, accurate and robust anisotropic rotation averaging. The proposed ACD solver is the simple extension to the family of block coordinate descent methods tackling the isotropic objective. It is the first dedicated solver that optimizes anisotropic chordal distances. We show that ACD has a large improvement in runtime compared to the state-of-the-art anisotropic SDP solver while delivering as accurate solutions.

\section{Acknowledgments}
This work was supported by the Wallenberg Artificial Intelligence, Autonomous Systems and Software Program (WASP), 
funded by the Knut and Alice Wallenberg Foundation, 
and the Swedish Research Council (grant no. 2023-05341). 

\bibliography{main}

\setcounter{equation}{0}
\setcounter{table}{0}
\setcounter{figure}{0}
\renewcommand{\theequation}{\thesection.\arabic{equation}}
\renewcommand\thefigure{\thesection.\arabic{figure}}
\renewcommand\thetable{\thesection.\arabic{table}}
\clearpage
\setcounter{page}{1}
{
\centering
\Large
\textbf{Making Rotation Averaging Fast and Robust with Anisotropic Coordinate Descent}\\
\vspace{0.5em}Supplementary Material \\
\vspace{1.0em}
}
\appendix

\section{Results on Challenging Data}
In Figures \ref{fig:qualitative1} and \ref{fig:qualitative2}, we show the estimated absolute rotations from much more challenging relative rotation inputs extracted from ETH3D MVS (DSLR) datasets~\cite{schops2017multi}. We run the two-view reconstruction pipeline on SIFT keypoints with minimal filtering; specifically, we filter matches only based on the cheirality condition. We use ground-truth camera translations and pre-align estimated absolute rotations with the ground-truth rotations robustly using RANSAC and $20^\circ$ threshold. ACD\textsubscript{\small AIRLS} often gives qualitatively the best results. Overall, both ACD and ACD\textsubscript{\small AIRLS} produce the fewest grossly incorrect rotations.

\begin{figure}[h]
\centering
\small
\caption{Qualitative rotation averaging results. Deviations from the ground truth are highlighted in red. Reported are RMS angular error and average accuracy.}
\vspace{5pt}
\begin{tabularx}{\textwidth}{CCCC}
    RCD~\cite{parra2021} & L1IRLS~\cite{chatterjee2013efficient} & \textbf{ACD} & \textbf{ACD\textsubscript{\small AIRLS}}
\end{tabularx}
\includegraphics[width=\textwidth]{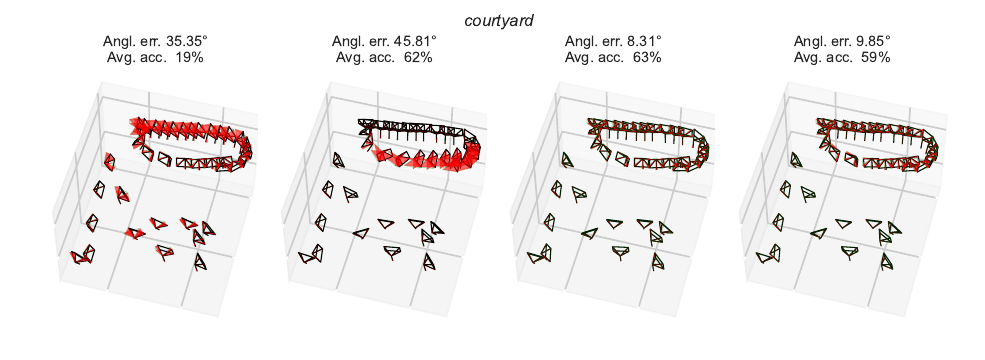}\\
\includegraphics[width=\textwidth]{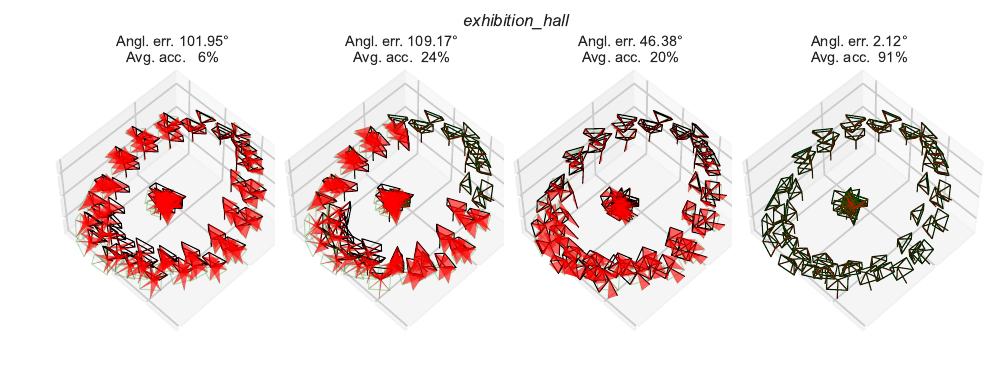}
\label{fig:qualitative1}
\end{figure}

\begin{figure}[h]
\centering
\small
\caption{(cont.) Qualitative rotation averaging results. Deviations from the ground truth are highlighted in red. Reported are RMS angular error and average accuracy.}
\vspace{5pt}
\begin{tabularx}{\textwidth}{CCCC}
    RCD~\cite{parra2021} & L1IRLS~\cite{chatterjee2013efficient} & \textbf{ACD} & \textbf{ACD\textsubscript{\small AIRLS}} 
\end{tabularx}
\includegraphics[width=\textwidth]{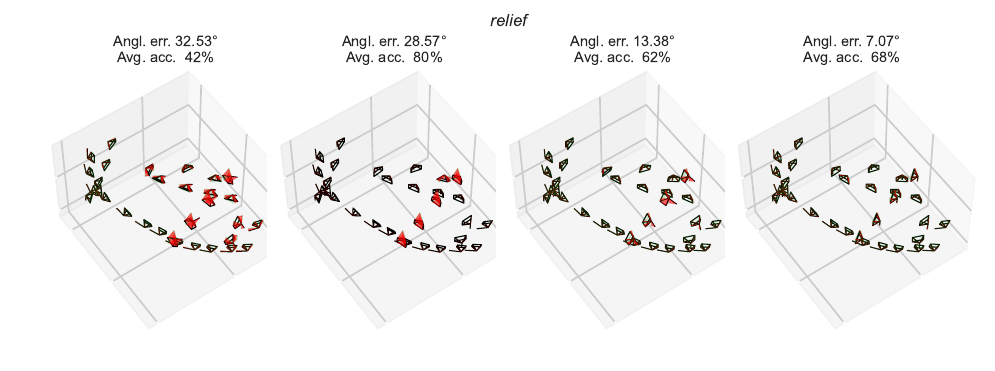}\\[-1em]
\includegraphics[width=\textwidth]{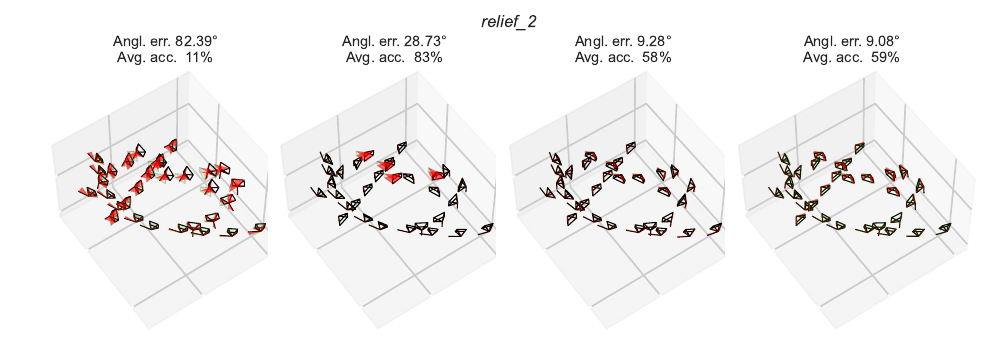}\\[-1em]
\includegraphics[width=\textwidth]{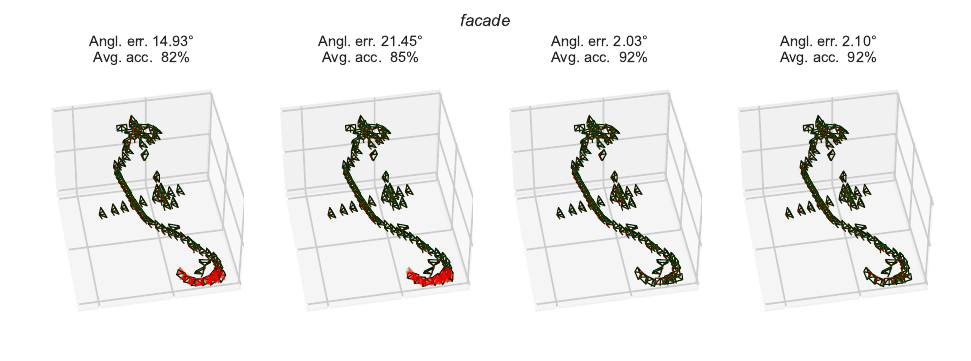}\\[-1em]
\includegraphics[width=\textwidth]{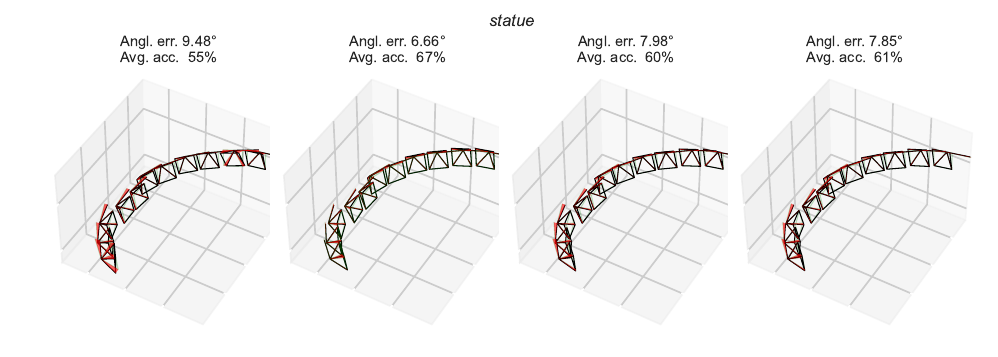}
\label{fig:qualitative2}
\end{figure}

\end{document}